\documentclass[10pt,twocolumn,letterpaper]{article}

\usepackage{authblk}
\usepackage{wacv}              % To produce the CAMERA-READY version
\usepackage{times}
\usepackage{epsfig}
\usepackage{graphicx}
\usepackage{amsmath}
\usepackage{amssymb}
\usepackage{enumitem}
\usepackage{multirow}
\usepackage{stmaryrd} 
\usepackage[dvipsnames]{xcolor}

\usepackage[breaklinks=true,bookmarks=false]{hyperref}

\definecolor{TE}{RGB}{204, 255, 204} %{218, 232, 252}
\definecolor{OD}{RGB}{204, 229, 255} %{218, 232, 252}
\definecolor{MD}{RGB}{255, 204, 204} %{255, 242, 204}
\definecolor{FT}{RGB}{255, 229, 204} %{225, 213, 231} 
\begin{document}

%%%%%%%%% TITLE
% \title{A Guide to Action Anticipation - Leveraging next-active-objects for Egocentric Videos \cig{NAOGAT: Next-Active-Object Guided Action Anticipation Transformer}}
\title{Leveraging Next-Active Objects for Context-Aware Anticipation \\ in Egocentric Videos}

\author[1,4]{Sanket Thakur}
\author[2, 1]{Cigdem Beyan}
\author[1]{Pietro Morerio}
\author[3, 1]{Vittorio Murino}
\author[1]{Alessio {Del Bue}}
\affil[1]{Pattern Analysis and Computer Vision (PAVIS), Istituto Italiano di Tecnologia (IIT)}
\affil[2]{University of Bergamo, Dalmine, Italy}
\affil[3]{University of Verona, Italy}
\affil[4]{University of Genoa, Italy}
\maketitle
\thispagestyle{empty}

%%%%%%%%% ABSTRACT
\begin{abstract}
   Objects are crucial for understanding human-object interactions. 
   By identifying the relevant objects, one can also predict potential future interactions or actions that may occur with these objects. In this paper, we study the problem of Short-Term Object interaction anticipation (STA) and propose NAOGAT (Next-Active-Object Guided Anticipation Transformer), a multi-modal end-to-end transformer network, that attends to objects in observed frames in order to anticipate the next-active-object (NAO) and, eventually, to guide the model to predict context-aware future actions.
   The task is challenging since it requires anticipating future action along with the object with which the action occurs and the time after which the interaction will begin, a.k.a. the time to contact (TTC). Compared to existing video modeling architectures for action anticipation, NAOGAT captures the relationship between objects and the global scene context in order to predict detections for the next active object and anticipate relevant future actions given these detections, leveraging the objects' dynamics to improve accuracy. One of the key strengths of our approach, in fact, is its ability to exploit the motion dynamics of objects within a given clip
   % This includes incorporating background motion information as object trajectories
   , which is often ignored by other models, and separately decoding the object-centric and motion-centric information. Through our experiments, we show that our model outperforms existing methods on two separate datasets, 
   Ego4D and EpicKitchens-100 (``Unseen Set''), as measured by several additional metrics, such as time to contact, and next-active-object localization. The code will be available upon acceptance.
\end{abstract}

\vspace{-10pt}
%%%%%%%%% BODY TEXT
\section{Introduction}

% "NAO: Your One-Stop Shop for Predicting Future Egocentric Actions"

\begin{figure}[t!]
\centering
\includegraphics[width=\linewidth]{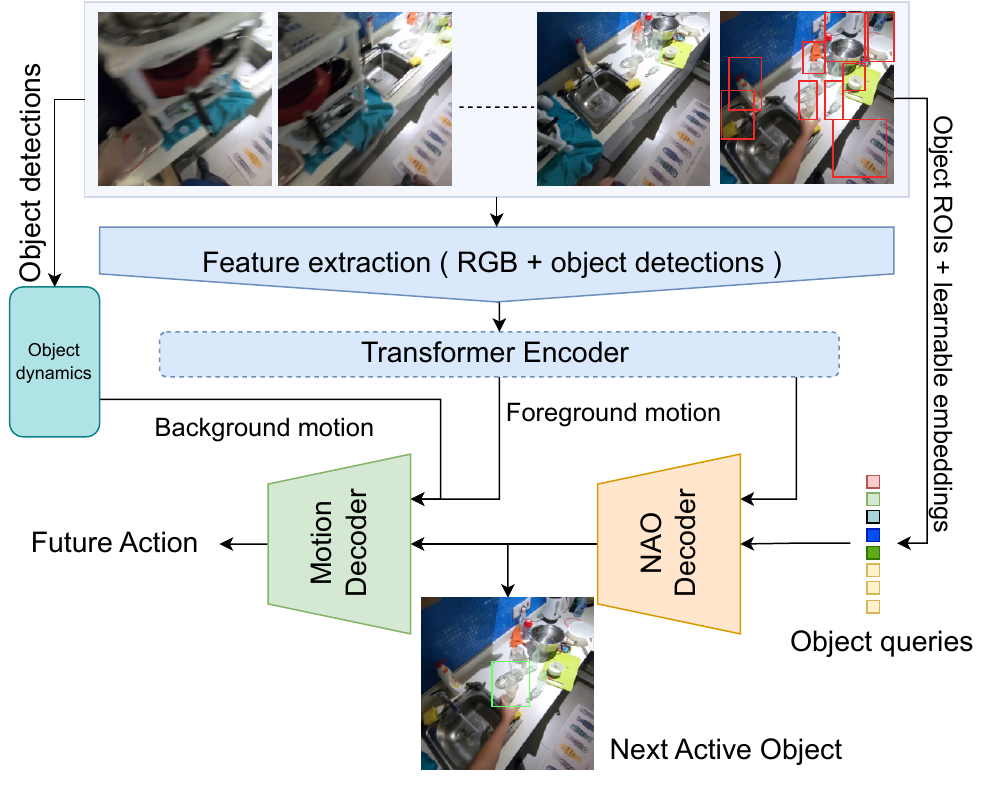}
\vspace{-10pt}
\caption{Our proposed model, NAOGAT, uses both features from video frames and object detections, which are combined and fed to a transformer encoder. Given object queries and encoded frame features from last observed frame, NAO decoder predicts relevant next-active-object detections. 
This information from NAO decoder is then passed along to the Motion decoder, which utilizes object dynamics to extract background information as object trajectories, to anticipate future frame interaction with encoder frame features (foreground motion) and next-active-object decoded features.
% object dynamics, the model predicts future actions. \VM{pretty clear and easy to follow :) @Sanket Try to read it yourself}
}
\label{fig:teaser}
\vspace{-15pt}
\end{figure}

Have you ever wondered how humans are able to effortlessly navigate their surroundings and perform actions based on what they see, especially in virtual reality (VR) and augmented reality (AR) environments ?  Such actions often involve contact with objects which are referred to as \textit{active objects} in the egocentric vision literature \cite{ADL}. Understanding and predicting the interactions with these objects are essential for enhancing the realism and interactivity of VR and AR experiences \cite{vr_obj_man, ar_man}.

For example, Fig. \ref{fig:teaser} shows a first-person video clip where someone is about to perform a specific action. % Considering an action that is to be performed as shown in Fig. \ref{fig:teaser}, %based on the observation of the person 
From the observed frames, it is reasonable to guess that the person is going to make a contact with the glass to possibly perform a \textit{wash} or \textit{fill} action. This reasoning helps us to recognize two important cues from a video: (1) which object will be ``used'' (i.e., active) in the future, and (2) what possible actions can be performed with that object. The category of objects significantly influences the nature of actions performed on them \cite{borghi2005object}. 
For example, a \textit{cut} action might not be performed in this scenario if we know that \textit{glass} is the next-active-object. 
Thus, intuitively, a model can make better predictions if it is able to anticipate which object(s) in the scene is possibly engaged in the very next future,  to support and drive the identification of the future action. This can enable more immersive and interactive experiences, where users can seamlessly interact with virtual objects based on anticipated actions, leading to enhanced user engagement and satisfaction.

This concept is particularly relevant in the Short-Term Anticipation (STA) task, which involves predicting the next-active-object (NAO) and its position, along with the time to contact (TTC) with that object, as well as the upcoming action, for a given video clip. 
This task depends on the assumption that the NAO is visible or present in the last observed frame, enabling its identification and localization \cite{ego4d}. The task that is more frequently exploited in the egocentric vision literature is instead action anticipation \cite{memvit2022,avt}, which refers to predicting a future action involving an object interaction without necessarily requiring the object to be visible in the last observed frame. 

The use of object-centric cues has shown significant promise in various video understanding tasks, such as action recognition \cite{orvit,obj_emb_behaviour,obj_AR}, hand-object forecasting \cite{hand_obj_joint,ego_obj_graph,obj1, obj_handobj}, and action anticipation \cite{rulstm,meccano,liu2019forecasting}. However, egocentric action anticipation methods \cite{avt,rulstm,liu2019forecasting,memvit2022} often overlooked such cues and mostly relied on holistic scene features and/or hand features. 
Indeed for STA, there exists no method explicitly considering %the notion of 
the information that could be gained from the object and more importantly NAOs.

In this paper, we propose a novel multi-modal architecture, called NAOGAT (Next-Active-Object Guided Attention Transformer), that involves training a model to attend to the objects in the last observed frame based on an obsered video clip, allowing it to predict the next-active-object (NAO). By incorporating the most relevant objects in the upcoming action and modeling the object dynamics of an observed clip, our model can make more accurate predictions of future actions and time to contact with those object(s) (NAO) and improve its overall performance.

The experimental analysis performed on two large-scale datasets: Ego4D \cite{ego4d} and EpicKitchen-100 (EK-100) \cite{ek100} demonstrate the favorable performance of NAOGAT with respect to several other methods, and indeed prove the importance of NAO cues and object dynamics for the targeted task. Particularly, on the Ego4D dataset \cite{ego4d}, we show a $2.16\%$ gain in Average Precision for VERB $+$ NAO, while a notable improvement of $7.33\%$ is observed in estimating the TTC $+$ NAO. 

The contributions of this paper can be summarized as follows: 
\begin{itemize} [leftmargin=*]
 \setlength\itemsep{0em}
 \item We present a Transformer-based method, called NAOGAT, for STA task, which models next-active-object anticipation as a fixed-set prediction problem based on object queries. 
 \item We propose a joint learning strategy based on fixed and learnable object queries which are extracted as ROIs from object detections and learned based on global context of video respectively: relevant detections for next-active-object guide the model to anticipate for object-specific future actions.
 \item The proposed method also exploit the motion dynamics of objects in sampled frames to model background information, in terms of object trajectories, along with foreground motion extracted from RGB features to better represent the human-object interaction.
\item We provide next-active-object annotations for the EpicKitchen-100 \cite{ek100} dataset in terms of NAO location \textit{w.r.t} last observed frame, which can be used to further advance research in egocentric video analysis and action anticipation.
\end{itemize}
%\VM{contributions above requires to quote also results, in the last point. AND the previous 3 points require better focus on what is really important and novel to highlight: as it's written now, I can't clearly distinguish the actual important contributions}

% \VM{AND the previous 3 points require better focus on what is really important and novel to highlight: as it's written now, I can't clearly distinguish the actual important contributions. WHAT'S IMPORTANT and NOVEL: transformer? NAO cues? Object motion dynamics and foreground motion (are they different)? the training method/the architecture? Again background and foreground motion dynamics??? I went through the contributions written in the first 3 points but I can't clearly catch which is important and why !}

\begin{figure*}[ht!]
\centering
\includegraphics[width=\linewidth, height=0.70\linewidth]{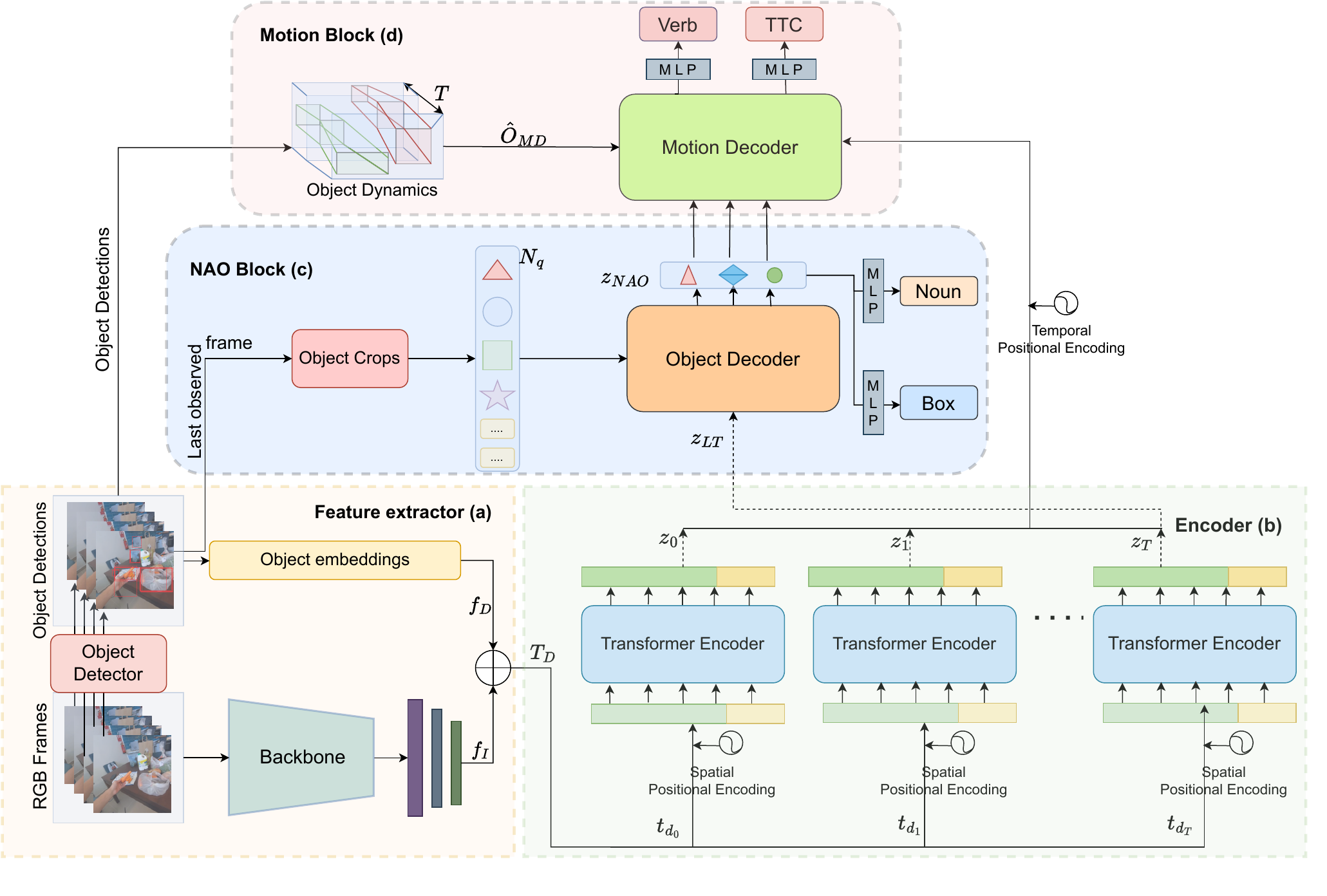}
\vspace{-25pt}
\caption{Our NAOGAT model first extracts feature information and object detections from a set of frames within an observed clip segment by means of a backbone network and an object detector; object detections are then transformed into object embeddings using an MLP network. The frame features are then concatenated with object embeddings to be sent to the transformer encoder after appending with spatial positional encoding. The encoder then extracts foreground motion (video memory) and global context features, which are used in two separate decoders to perform object-centric and motion-centric predictions. For object decoder, detections from the last observed frame and learnable embeddings are used to create object queries to perform fixed-set predictions for the NAO class label and its bounding box using a transformer decoder. In the last stage, we leverage Object dynamics to extract background motion in terms of object trajectories for detected objetcs in sampled frames. We then use the object decoder's outputs with the combined frame representation of video memory and object dynamics to perform predictions for motion-related outputs, such as future action and time to contact (TTC).
} 
\centering
\label{fig:method}
\vspace{-10pt}
\end{figure*}

\section{Related Work}
Existing action anticipation methods in egocentric videos have primarily focused on utilizing features extracted from video clips, but they often overlooked the importance of objects and their interactions. Below, we discuss the existing works on the next-active-object and overall action anticipation methods in first-person videos since in this paper we aim to show that leveraging the next-active-objects would improve the action anticipation task.

\noindent \textbf{Next-active-object.}
In egocentric vision literature, \emph{active} and \emph{passive} objects were first time introduced by Pirsiavash and Ramanan \cite{ADL}, which defined the active objects as those the first-person is interacting with, while passive objects stand for the contrary. Dessalene et al. \cite{tpami_contact} used the \cite{ADL} description and proposed a method to predict \emph{next-active-objects}, however, they limited their objective only to the objects that are contacted by the first-person's \emph{hand}. Consequently, this approach requires the hands and the next-active-objects to be visible in the frames, which might be a significant limitation in practical applications. 
Furnari et al. \cite{furnari2017next} utilized object tracking to detect the next-active-objects, limited to being able to predict them only in \emph{one} future frame. 
Other works \cite{hand_obj_joint,JIANG2021212} proposed to identify the objects either in a single image (without analyzing the spatiotemporal data) or by predicting the future hand motion. Both fall behind in explicitly exploiting the future active object information to predict future action. Recently, Thakur et al. \cite{anacto} proposed a transformer-based approach that applies collaborative modeling of RGB and object features to anticipate the location of the next-active-object(s) several frames ahead of the last observed frame. 

\noindent \textbf{Action Anticipation.}
This task stands for forecasting the future actions of a person given an egocentric video clip including the past and current frames. Several approaches in this line have focused on learning the scene features with Convolutional Neural Networks (CNNs), e.g., by  modeling the hand-object contact points \cite{liu2019forecasting}. Others aggregated the past contextual features \cite{rulstm, feichtenhofer2019slowfast, sener2020temporal}, and a few focused on modeling the future interaction of consecutive frames \cite{future_frame}. With the emergence of Vision Transformers \cite{vit,video_swin}, researchers have started to investigate 
the utility of transformers in their work. For instance, \cite{avt} proposed causal modeling of video features, introducing sequence modeling of frame features to decode the interactions in consecutive future frames. We et al. \cite{memvit2022} proposed a long-term understanding of videos using Multiscale Transformers by hierarchically attending to previously cached memories. In a recent work, Zhang et al. \cite{Zhong_2023_WACV} proposed fusing object information along with RGB frame features for a better video context representation in addition to investigating the impact of the
audio modality for action anticipation.
Our work distinguishes itself from the related work by addressing action anticipation in egocentric videos with the prediction of the next-active-objects. 

%In this work, we leverage the object features extracted from object detectors \cite{fastercnn,ek55} to provide additional priors for the action anticipation task. To this extent, we separately predict the future ``active objects'' to refine the prediction of the future action, that is to be performed by a person.
%In detail, unlike the aforementioned studies, our work builds on the concept of the next-active-object prediction by leveraging information from detected objects in a given clip. 
%Our proposed method not only predicts the next active objects but also identifies the objects that may not be detected by a pre-trained object detector. This enables the prediction of a wider range of possible next-active objects for future actions.

\section{Method}
\label{sec:method}

As already mentioned, the Short-Term Anticipation (STA) task aims at predicting the next human-object interaction happening after a certain (unknown) time $\delta$, named the Time To Contact (TTC), relying on evidence up to time $T$. Thus our model's input is a $T$-frames video sequence $V = \{{v_i\}^T_{i=1}}$, where $v_i \in \mathbb{R}^{C\times{H_o}\times{W_o}}$ is an RGB frame, while it must output 4 unknowns at time $T$, following the protocol introduced in ~\cite{ego4d}: 
% the position of the next-active-object, the object name, the action verb, and the very same TTC $\delta$. %in the last observed frame and anticipate the future interaction with that object in $\delta$ seconds, where $\delta$ is unknown. 
% Our NAO outputs a fixed set of predictions, where each prediction indicates the next-active-object in terms of 
the NAO noun class ($\hat{n}$), its location, i.e. bounding box ($\hat{b}$), a verb depicting the future action ($\hat{v}$) and the time to contact ($\hat{\delta}$), which estimates how many seconds in the future the interaction with the object will begin. \\
% It is imperative that for this task, the next-active-object is visible in the last observed frame \cite{ego4d}, and the model is required to make predictions at a specific timestamp \textit{i.e;} frame $T$, rather than densely throughout the video. 

\subsection{Method Overview}

% \\

% [FIXME: What is happening here? There are 4 paragraphs with no apparent link between them! HERE you have to do an introduction of the model following Figure 2. If you can do this well, the paper will be rejected.]

% \\
We now introduce our model architecture, as illustrated in Fig. \ref{fig:method}, which is designed to predict the class and location of the next-active-object and the time required to make contact with the object (the TTC $\delta$) as well as the future action, based on a given observed clip. The proposed method comprises 4 main modules. First, a \emph{feature extractor} (Fig. \ref{fig:method}\colorbox{FT}{(a)}, Sec. \ref{subsec:feat}) operates on frames from a sampled video clip to extract RGB and object detection features. This is followed by an \emph{Encoder Block} (Fig. \ref{fig:method}\colorbox{TE}{(b)}, Sec. \ref{subsec:enc}) that operates on the combined features to facilitate the exchange of information across frames. Following the encoder, the two separate head architectures - \emph{NAO Block} (Fig. \ref{fig:method} \colorbox{OD}{(c)}, Sec. \ref{subsec:nao}) and \emph{Motion Block} (Fig. \ref{fig:method}\colorbox{MD}{(d)}, Sec. \ref{subsec:motion}), is used to predict the next-active-object information and future action prediction respectively. Our model employs object queries from object detection in the last observed frame to locate and identify the next-active-object and is inspired by the direct set prediction problem \cite{detr}, which involves predicting a fixed set of objects and modeling their relationship. In addition, we leverage object dynamics to extract background motion in a video clip. This involves incorporating object trajectories for detected objects in sampled frames within the \emph{Motion Block} and utilizing attended frame features from the encoder module to model the relationship with NAO priors (Fig. \ref{fig:method}-(c)), in order to predict future action and TTC. An overview of our model architecture in shown in Fig. \ref{fig:teaser}, which is described in detail in Fig. \ref{fig:method}. In the following we describe each model component in detail, followed by training and implementation details.

\subsection{Feature extractor}
\label{subsec:feat}
For a given video clip, we sample a set of $T$ frames $V = \{v_i\}^T_{i=1}$, where, $v_i \in \mathbb{R}^{C\times{H_o}\times{W_o}}$, which are fed to \textit{i)} a backbone network for feature extraction \textit{ii)} a pre-trained object detector (Fig. \ref{fig:method}-(a)).

\textit{i)} While a number of video-based backbone architectures have been proposed \cite{orvit,memvit2022,vit,video_swin,i3d} to extract frame-level feature representation from a given video clip, for the task of STA a suitable spatial-temporal encoder is required to be able to extract both static appearance information (\textit{e.g.,}, object location, size) and motion cues. Video Swin Transformer \cite{video_swin}, a recently proposed spatial-temporal transformer architecture was proposed based on shifted windows architecture of Swin Transformer \cite{swin} for the video domain. Video Swin contains just a single temporal downsampling layer and can be easily adapted to output per-frame feature maps, essential for us to localize and identify the next-active-objects at the same time reasoning on the whole sequence. Specifically, we adopted the Swin-T \cite{video_swin} architecture as our backbone.  The frames $V$ are given in parallel to the video swin architecture to extract frame features, $f_I \in \mathbb{R}^{T \times H \times W \times C^{'} } $, where $T$, $H$ and $W$ denotes the temporal length of video clip, as well as height and width of the feature maps respectively. 

\textit{ii)} In addition, a Faster R-CNN \cite{fastercnn} based object detector pre-trained on Ego4D\cite{ego4d}, is used to extract the object detections from the sampled frames, in terms of bounding boxes coordinates and confidence score, resulting in a $(4+1)$-dimensional vector. We limit the number of bounding boxes to be used to a fixed number $Q$ to maintain a consistent number of detections across each frame. If there are fewer detections than $Q$, then dummy coordinate and score values corresponding to no detection are appended. 
Finally an MLP processes $5$-d vectors into object embeddings $f_D \in \mathbb{R}^{T \times Q\times{D}}$. 

Finally, visual features are also projected to a shared dimension $D$ as of $f_D$, using a 2D convolution layer with kernel size as 1, $f_I \in \mathbb{R}^{T \times H \times W \times D }$. Finally, the features from each modality are then flattened and \textit{separately} concatenated along the temporal dimension, producing a set of $T_D = \{{{t_d}_i\}^T_{i=1}}$ multimodal embeddings, where ${t_d}_i \in \mathbb{R}^{(H\times{W} + Q)\times{}D}$.

\subsection{Transformer Encoder}
\label{subsec:enc}
In the next step, according to Fig. \ref{fig:method}-(b), the concatenated multimodal embeddings, $T_D$ are simultaneously passed to a Transformer Encoder \cite{detr} after appending spatial positional encoding. The Transformer Encoder blocks allow exchanging frame-level information within inter-frame features while maintaining the same dimension. The output of the encoder is $Z$, where $Z \in \mathbb{R}^{T\times(H\times{W} + Q)\times{}D}$,  which is the combined features representation across frames and object detections. It is split into 2 parts : 
1) Global-context memory, $z_{LT}$, where $z_{LT} \in \mathbb{R}^{(H\times{W} + Q)\times{}D}$ extracted from last frame of $Z$, 
2) Video-only memory: $\{{z_i\}^T_{i=1}}$  where $z_i \in \mathbb{R}^{H\times{W} \times{}D}$, which aims to captures foreground motion cues such as \textit{e.g., hands in first-person vision} \cite{avt}. The Global-context memory and Video-only memory are then used by the NAO and Motion blocks to find the instances that corresponds to possible next-active-object and also anticipate the future action respectively.

\subsection{NAO Block }
\label{subsec:nao}
The STC task requires anticipating the location of the next-active-object wrt the \textit{last} frame observed by the model. Therefore, as shown in Fig. \ref{fig:method}-(c), we only use the features corresponding to the last frame from the transformer encoder, namely $z_T$, as input to the NAO block, along with $N_q$ object queries for the frame of interest. We define our object queries as the regions of interest (ROIs) extracted by the object detector, i.e., a feature map for each detection. If there are no sufficient detections, \textit{i.e.,} the number of detections is less than $N_q$, then we append learnable tokens for the rest of the queries. Our object decoder follows the standard architecture of the transformer decoder \cite{detr}, transforming $N_q$ embeddings of size $D$ using multi-headed attention mechanisms. The $N_q$ object queries are decoded by using $z_T$ as key/value pairs in the multiple multi-head attention layers. The decoded features, $z_{NAO} \in \mathbb{R}^{N_q \times D}$, are then used to predict bounding box coordinates ($\hat{b}$) and class labels ($\hat{n}$) by an additional MLP block, resulting in $N_q$ final predictions for the next-active-object. The decoder's primary function is to attend to objects ( detected/learned ) in the last observed frame based on a global context of a video clip, resulting in the prediction of a possible next-act-object and its corresponding object label.

\subsection{Motion Block }
\paragraph{\textbf{Object Dynamics.}}
\label{subsec:motion}
We propose to integrate the video frame features from the transformer encoder with the object dynamics of detected objects in the video clip, in order to better estimate the time required to approach the next-active-object predicted by our object decoder (Sec. \ref{subsec:nao}). As shown in Fig. \ref{fig:method}-(d), object dynamics refer to a proxy for object trajectories of background objects in the video clip. \cite{orvit} previously used object dynamics to enhance the frame representation for effective motion information modeling in videos. However, their approach requires object region information and multiple stacking of the feature representation block in a transformer encoder for action recognition benchmark. In contrast, we treat object motion dynamics as a separate module for extracting object traversals. Object trajectories are the bounding box movement across the frames in sampled video clip. 
We interpret these trajectories as background motion because they correspond to passive motion in the scene, and combine them with transformer encoder outputs to model human-object motion features. This approach has proven useful in estimating the speed of interaction and predicting motion-centric information.
The Object Dynamics block takes as input the object detection's box locations $od_i$ for a frame $i$ and outputs spatial-temporal tokens, 
\begin{equation}
   \hat{o_{MD}}_0, \dots \hat{o_{MD}}_T = \textit{OMD}(od_0, \dots , od_{T})
\end{equation}
where $\hat{o_{MD}}_i \in \mathbb{R}^{H\times W \times D}$.
In $OMD$, initially, each object detection is expanded from $T\times Q\times4$ into $T\times Q\times{D}$ tokens using a MLP. These tokens are flattened then used to perform self-attention operation and projected on a spatial-temporal dimension $THW\times{D}$ using a bi-linear interpolation sampler operation \cite{MD_1} to output object trajectories for frames used in the inputs $\hat{o_{MD}}_i$.
 %\PM{Sanket, explain how the module is actually working! i.e. more details on OMD. I guess it is a simple self attention layer, give some more details, possible a formula.}   
This module provides more detailed information on the object motion, \textit{i.e.;} background motion of frames.

\paragraph{Motion Decoder.}
\label{sec:motio_dec}
It was empirically observed that a single Object Decoder (Sec. \ref{subsec:nao}) leads to the dropping of motion information across the frames, resulting in very poor performance for future action prediction. For this purpose, we decided to use a separate decoder for motion-related predictions (verb and TTC). Inspired from \cite{avt}, we additionally combine frame features with the object motion dynamics features to model the foreground motion from video memory (Sec. \ref{subsec:enc}) and background motion from object dynamics (Sec. \ref{subsec:motion}) at the frame level.
\begin{equation}
  \small{  z'_0, \dots z'_T = MLP(LN(z_0, \dots , z_{T} \bigoplus \hat{o_{MD}}_0, \dots , \hat{o_{MD}}_{T})) }
\end{equation}
 Here, object motion dynamics features, $\hat{o_{MD}}_i$ are added to encoder features $z_i$ along spatial and temporal dimension \textit{T, H, W}, where $\bigoplus$ denotes such element-wise summation. This is followed by a Layer Norm (LN) and a MLP. 
 In addition, to influence our future action prediction based on our next-active-object prediction, we add the object decoder embeddings, $z_{NAO}$ to the last observed frame before feeding the sequence to the decoder. 
 \begin{equation}
     \hat{z_1}, \hat{z_2}, \dots , \hat{z_{T+1}} = \textit{D}(z'_0, z'_1, \dots , z'_T + z_{NAO})
 \end{equation}
 We implement $D$ using the masked transformer decoder as followed in popular approaches such as \cite{gpt2}. We feed the modified inputs features to the masked decoder after appending with temporal positional encoding. The masking ensures that the model attends to specific parts of the input while performing the prediction for the next consecutive position. That helps our model to understand the interaction of person and the surrounding motion. The additional input of $z_{NAO}$ helps to refine future action prediction. The design differs considerably from \cite{avt}, since we model the background and foreground motion in a combined fashion with additional priors of next-active-object added to last observed frame features before the causal modeling. The decoder network $D$ is designed to produce attentive features corresponding to the future frames using the object motion dynamics and also the next-active-object information in the last observed frame to anticipate the future action. 
We use the future frame feature, $z_{T+1}$  to predict future action label $\hat{v}$ and the TTC $\delta$ corresponding with the next-active-object obtained from the object detector, using a feed-forward network.

\subsection{Training}
Let us denote $y$ as the set of ground truth set of objects, and $\hat{y} = \{{{\hat{y}}_i\}^N_{i=1}}$ as the set of $N$ predictions, relating to $N$ object queries. Based on the procedure of finding matching elements of \cite{detr}, we identify the one-to-one matching for the predictions with the ground truth labels using the \textit{Hungarian loss} for all pairs matched.

\noindent \textbf{Bounding box loss.}
The major difference between us and \cite{detr} is that we aim to learn bounding boxes based on some initial guesses, rather than only performing the predictions directly. The predicted bounding boxes are regressed using a combination of L1 loss and the generalized IoU loss and are defined as : 
\begin{equation}
  \small{  \mathcal{L}_{box} = \lambda_{iou}\mathcal{L}_{iou}(b_i, \hat{b}_{\sigma(i)}) + \lambda_{L1}||b_i - \hat{b}_{\sigma(i)}||_1
    \label{eq_bbox} }
    \vspace{-2pt}
\end{equation}
where $\lambda_{iou}, \lambda_{L1} \in \mathbb{R}$ are hyper-parameters.

\noindent \textbf{Classification losses.}
The second loss, denoted by $\mathcal{L}_{noun}$ and $\mathcal{L}_{verb}$, is a cross-entropy loss that supervises the prediction of labels for the next-active-object and the future action: 
\begin{equation}
    % \mathcal{L}_{noun}(\hat{y_i}, y_i), 
    \mathcal{L}_{verb/noun}(\hat{y_i}, y_i) = {\sum_{t=0}^{N}{y^t_i}.\log(\hat{y}^t_i)}
    \label{eq_class}
    \vspace{-2pt}
\end{equation}
% \begin{align}
%     \mathcal{L}_{noun}(\hat{y_i}, y_i) = {\sum_{t=0}^{N}{y^t_i}.\log(\hat{y}^t_i)} \\ 
%     \mathcal{L}_{verb}(\hat{y_i}, y_i) = {\sum_{t=0}^{N}{y^t_i}.\log(\hat{y}^t_i)}
%     \label{eq_class}
%     \vspace{-5pt}
% \end{align}

\noindent \textbf{Regression and feature Loss.}
The regression loss, denoted by $\mathcal{L}_{ttc}$ is the smooth L1 loss \cite{smoothl1} and is used to train the model to regress the time to contact prediction. Finally, we also use a feature loss, ${L}_{feat}$ defined below in Eq. \ref{eq_lfeat} from \cite{avt} which aims at leveraging the predictive structure of the motion decoder \ref{sec:motio_dec}: the decoder is basically trained to predict future frame features given frames up to time $t$ only. 
%by supervising the future frame features predicted by the decoder to match the true future frame features that are extracted as embeddings from the encoder. %\vspace{-0.2cm}
\begin{equation}
    \mathcal{L}_{feat} = \sum_{t=0}^{N}||\hat{z}_{t+1} - z'_{t+1}||^2_2,
    \label{eq_lfeat}
    \vspace{-5pt}
\end{equation}
% \begin{equation}
%     \mathcal{L}_{noun}(\hat{y_i}, y_i) = \lambda_{c}{\sum_{t=0}^{N}{y^t_i}.\log(\hat{y}^t_i)}
%     \label{eq_class}
%     % \vspace{-5pt}
% \end{equation}

In the end, all losses are combined to produce the overall loss:
\begin{equation}
    \mathcal{L} = \mathcal{L}_{box} + \lambda_{2}\mathcal{L}_{noun} + \lambda_{3}\mathcal{L}_{verb} + \lambda_{4}\mathcal{L}_{ttc} + \mathcal{L}_{feat}
    \label{eq_overall}
    \vspace{-5pt}
\end{equation}
where $\lambda_{2}, \lambda_{3}, \lambda_{4} \in \mathbb{R}$ are hyperparameters.

\begin{table*}[t]
% \vspace{-5pt}
\begin{center}
\resizebox{\linewidth}{!}{
\begin{tabular}{|l||cccccccc|} \hline
Models & $AP_{\hat{b}}$ & $AP_{\hat{b} + \hat{n}}$ & $AP_{\hat{b} + \hat{n} + \delta}$ & $AP_{\hat{b} + \hat{n} + \hat{v}}$ & $AP_{\hat{b} + \hat{n} + \hat{v} + \delta}$ & $AP_{\hat{b} +  \delta}$ & $AP_{\hat{b} + \hat{v}}$ & $AP_{\hat{b} + \hat{v} + \delta}$  \\  \hline\hline
Slowfast \cite{ego4d} & 40.5 & 24.5 & 5.0 & 4.9 & 1.5 & 8.4 & 8.16 & 1.9  \\
Slowfast (with Transformer backbone) & 40.5 & 24.5 & 4.5 & 4.37 & 1.73 & 7.5 & 8.2 & 1.3 \\ 
AVT \cite{avt} & 40.5 & 24.5 & 4.39 & 4.52 & 1.71 & 7.12 & 8.45 & 1.15 \\
ANACTO \cite{anacto} & 40.5 & 24.5 & 4.55 & 5.1 & 1.91 & 7.47 & 8.9 & 1.54 \\
MeMVIT \cite{memvit2022} & 40.5 & 24.5 & 4.95 & 5.89 & 1.34 & 9.27 & 10.04 & 2.11 \\ \hline 
Ours  & \textbf{45.3} & \textbf{27.0} & \textbf{9.0} & \textbf{6.54} & \textbf{2.47} & \textbf{16.6} & \textbf{12.2} & \textbf{4.18} \\\hline
\end{tabular}}
\end{center}
\vspace{-10pt}
\caption{Results of our model and other baseline methods on Ego4D \cite{ego4d} dataset for different output targets, bounding box ($\hat{b}$), next-active-object label ($\hat{n}$), future action ($\hat{v}$) and the time to contact with the object ($\delta$) based on their Average Precision ($AP$). }
\label{table:avgpre}
\vspace{-10pt}
\end{table*}

% Article top matter
\begin{table*}[t]
\centering
\begin{center}
\begin{tabular}{@{\extracolsep{2pt}}|l|c|c||cccccc|ccc@{}|}
\hline
\multirow{2}{*}{Model} & Params & Init &\multicolumn{3}{c}{Unseen} &\multicolumn{3}{c}{Tail} &\multicolumn{3}{c|}{Overall} \\
\cline{4-6} \cline{7-9} \cline{10-12}
& (M) &  & Action & Verb & Noun & Action & Verb & Noun & Action & Verb & Noun$\;$ \\
\hline\hline
Chance & - & - & 0.5 & 14.4 & 2.9 & 0.1 & 1.6 & 0.2 & 0.2 & 6.4 & 2.0 \\
TempAgg (RGB) \cite{sener2020temporal} & - & \cite{in1k} & 12.2 & 27.0 & 23.0 & 10.4 & 16.2 & 22.9 & 13.0 & 24.2 & 29.8 \\
% \hline
RULSTM \cite{rulstm} & - & \cite{fastercnn} & - & - & - & - & - & - & 7.8 & 17.9 & 23.3 \\
RULSTM \cite{rulstm} & - & \cite{in1k} & 13.1 & 28.8 & 23.7 & 10.6 & 19.8 & 21.4 & 13.25 & 27.5 & 29.0 \\
% \hline
AVT (RGB) \cite{avt} & 393 & \cite{in1k} & - & - & - & - & - & - & 14.9 & \underline{30.2} & 31.7 \\
AVT + \cite{avt} & - & \cite{in1k} & 11.9 & \textbf{29.5} & 23.9 & \textbf{14.1} & \underline{21.1} & \underline{25.8} & \textbf{15.9} & 28.2 & \underline{32.0} \\ 
% \hline
MeMViT \cite{memvit2022} & 59 & \cite{k400} & 9.8 & 27.5 & 21.7 & \underline{13.2} & \textbf{26.3} & \textbf{27.4} & \underline{15.1} & \textbf{32.8} & \textbf{33.2} \\
\hline
Ours  & 23.5 & - & \textbf{14.3} & \underline{29.3} & \textbf{27.8} & 4.4 & 13.2 & 13.8 & 10.7 & 25.3 & 27.9 \\
% Ours (CNN backbone) & - & - & 8.9 & 19.7 & 22.3 & 8.5 & 25.5 & 19.8 & 3.2 & 9.9 & 10.1 \\
% \hline
\hline
\end{tabular}
\end{center}
\vspace{-10pt}
\caption{Results of our model and other baseline methods on EK-100 \cite{ek100} dataset on validation set. ``Overall'' comprises of samples combining the Unseen and Tail set plus also consisting of \emph{seen} samples from the training set. }
\label{tab:ek100_AA}
\vspace{-15pt}
\end{table*}

\begin{table*}[t]
% \vspace{-5pt}
\begin{center}
\resizebox{\linewidth}{!}{
\begin{tabular}{|l||cccccccc|} \hline
Model & $AP_{\hat{b}}$ & $AP_{\hat{b} + \hat{n}}$ & $AP_{\hat{b} + \hat{n} + \delta}$ & $AP_{\hat{b} + \hat{n} + \hat{v}}$ & $AP_{\hat{b} + \hat{n} + \hat{v} + \delta}$ & $AP_{\hat{b} +  \delta}$ & $AP_{\hat{b} + \hat{v}}$ & $AP_{\hat{b} + \hat{v} + \delta}$  \\  \hline\hline
Ours w/o \textit{OMD}, \textit{OD} & 45.3 & 26.7 & 4.78 & 5.55 & 1.05 & 7.89 & 8.91 & 1.52  \\
Ours  w/o  \textit{OMD} & 45.1 & 27.1 & 4.48 & 6.2 & 1.0 & 7.34 & 10.2 & 1.44  \\
Ours (ResNet50) & 42.7 & 25.2 & 4.3 & 6.0 & 1.1 & 10.6 & 10.1 & 2.0  \\
Ours (Full) & \textbf{45.3} & \textbf{27.0} & \textbf{9.0} & \textbf{6.54} & \textbf{2.47} & \textbf{16.6} & \textbf{12.2} & \textbf{4.18}  \\
\hline
\end{tabular}}
\end{center}
\vspace{-10pt}
\caption{Ablation study performed on ego4D \cite{ego4d} to investigate the effect of Backbone,  Motion Dynamics (MD), and  object decoder (OD) modules on the motion-based output sequences by the model.}
\label{table:ablation}
\vspace{-10pt}
\end{table*}

\begin{figure*}[t]
\includegraphics[width=\linewidth]{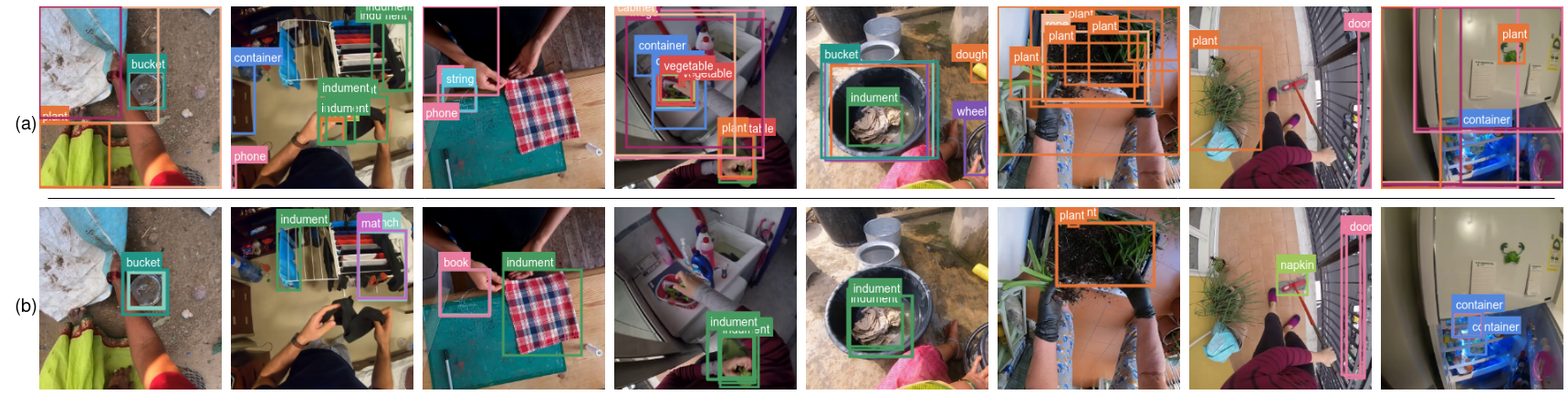}
\caption{The top row (a) shows the ``last observed frame" and all the object detections provided by the object detector \cite{fastercnn}. The bottom row (b) depicts the output from our motion decoder. It can be observed that our model learns from past observations and selects the best possible object(s) for the next-active-object selection in the frame. Besides, it can be seen that it is even able to identify objects which were not detected by the object detector (3$^{rd}$ and 7$^{th}$ column are the clearest examples). }
\vspace{-15pt}
\label{fig:qualitative}
\end{figure*}

\section{Experiments}
\subsection{Datasets}
%\noindent \textbf{Datasets}. 
\label{sec:datasets}
We used the following datasets to validate the effectiveness of our method quantitatively and qualitatively.

\noindent \textbf{Ego4D} \cite{ego4d} is currently the largest first-person dataset available, consisting of 5 splits covering distinct tasks and a total of 3,670 hours of videos across 74 different locations. For the next-active-object prediction and STA task, we use the ``forecasting split'' which contains over 1000 videos and is annotated at 30 fps for the STA task. The dataset annotations include the next-active-objects in the last observed frame, which is a unique feature of that dataset \textit{wrt.} STA. Our goal is to predict the noun class ($\hat{n}$), bounding box ($\hat{b}$), the verb depicting the future action ($\hat{v}$), and the Time to Contact (TTC) ($\delta$) for a given video clip. For comparison on the Ego4D dataset, we apply the existing methods, which are designed for action anticipation-based tasks, by confining them to only predict the next-active-object class label ($\hat{n}$), the verb depicting the future action ($\hat{v}$), and the TTC ($\delta$) since the compared methods are not designed to predict bounding boxes. \\

\noindent \textbf{Epic-Kitchens-100} \cite{ek100} consists of about 100 hours of recordings with over 20M frames comprising daily activities in kitchens, recorded with 37 participants. It includes 90K action segments, labeled with 97 verbs and 300 nouns (i.e. manipulated objects). Since the dataset does not provide annotation for next-active-object, we exploit the object detector provided by \cite{ek55} and also the annotations provided in \cite{anacto} to curate labelings composed of bounding boxes \textit{i.e;} locations of next-active-objects in the last observed frame. It is to be noted that, to adapt this dataset for the next-active-object detection task, it is imperative that the object, which is used in future action is visible in the last observed frame. However, based on our annotations we realized that for 12.5 $\%$ of training data in EK-100 the next-active-object annotations are absent \textit{i.e;} the future active object is not visible in the last observed frame.

\subsection{Implementation Details}
\label{sec:impdet}
In order to pre-process the input video clips, we randomly scale the height of the clips between 248 and 280 pixels and take 224-pixel crops for training. We sample 16 frames at 4 frames per second (FPS). We adopt the network architecture of Swin-T \cite{swin1,swin2} to serve as the backbone of our network to extract the video features from the sampled clip. However, we only utilize the outputs till the first-three block of the video swin transformer \cite{swin} along with down-sampling of each block to extract the \textit{per-frame} feature maps, which are required later to predict the bounding boxes. 
We also use a 3-layer multi-head transformer encoder and decoder, which operates on a fixed 256-D. We train our end-to-end model with SGD optimizer using a learning rate of $1e-4$ and a weight decay of $1e-6$ for 50 epochs. 

\subsection{Evaluation Metrics}
\label{sec:metric}
We evaluate our models on the Ego4D \cite{ego4d} dataset using the evaluation metrics defined by the dataset creators for short-term anticipation tasks. These metrics include the Average Precision of four different combinations of the next-active-object-related predictions: noun class ($\hat{n}$), bounding box ($\hat{b}$), future action ($\hat{v}$), and time to contact ($\delta$). We use the top-1 accuracy to evaluate the performance of the future action ($\hat{v}$) and next-active-object label ($\hat{n}$) predictions. For bounding boxes ($\hat{b}$) and time to contact ($\delta$), the predictions are considered correct if the predicted boxes have an Intersection over Union (IoU) value greater than or equal to 0.5 and the absolute difference between the predicted and ground-truth time to contact is less than or equal to 0.25 seconds ($|\hat{y}_{ttc} - y_{ttc}| \leq 0.25$). In the case of combined predictions involving two or more unknowns, the prediction is deemed correct only if all the unknowns are predicted correctly. For the purpose of training, we kept the values of 
all $\lambda$ as 1, except $\lambda_{4}$ which is set to 10 following \cite{ego4d}.

For comparison of models on the EK-100 \cite{ek100} dataset, we adhere to the metric commonly used in recent action-anticipation works \cite{rulstm,avt,memvit2022}.

% \begin{table}[h!]
% \begin{center}
% \resizebox{0.8\columnwidth}{!}{
% \begin{tabular}{|l||cccc|} \hline
% Model & BN& BNV & BNT & BNVT   \\ \hline\hline
% Baseline & 17.45 & 4.65 & 4.25 & 1.5  \\
% AVT & 17.45 & 0.8 & 3.89 & 0.18 \\ \hline
% Ours  & 17.8 & 5.75 & 6.6 & 2.3  \\\hline
% \end{tabular}}
% \end{center}
% \caption{Comparison between our model and baseline methods based on the metric defined above in Sec. \ref{sec:result}}
% \label{table:comp}
% % \vspace{-20pt}
% \end{table}

% \begin{table}[h!]
% \begin{center}
% \resizebox{\columnwidth}{!}{
% \begin{tabular}{|l||cccc|} \hline
% Model & Parameters (M) & Input frames & GLOPs & Frames   \\ \hline\hline
% Slowfast & 24.45 & 224 $\times$ 224 & - & 32 \\
% AVT & 393 & 224 $\times$ 224 & - & 10 \\
% MeMViT & 59.5 & 224 $\times$ 224 & - & 32\\ \hline
% Ours & 28.3 & 224 $\times$ 224 & - & 15 \\\hline
% \end{tabular}}
% \end{center}
% \caption{Comparison between our model and baseline methods based on model size.}
% \label{table:comp_size}
% % \vspace{-20pt}
% \end{table}

\subsection{Comparison with State-of-the-art}
For Ego4D dataset \cite{ego4d}, we compare our model with the methods restricted to only predicting the future action ($\hat{v}$) and TTC ($\delta$) of a given sample clip, since the only methods we perform a comparison with, are action anticipation methods that have not been designed to predict bounding boxes. Table \ref{table:avgpre} declares the results for Ego4D \cite{ego4d} dataset. We observe that our model achieves better performance than the object detector \cite{fastercnn} that is pre-trained on Ego4D, in terms of predicting the NAO's class label and bounding box location, as evidenced by the higher ${AP}_{\hat{b}}$ and ${AP}_{\hat{b} + \hat{n}}$ scores. This superiority is also visually evident in Fig. \ref{fig:qualitative}, where the performance of our object decoder is shown to refine the detected objects for NAO and even identify objects that were not detected by the object detector. Moreover, our model outperforms all the other baseline methods across all other evaluation metrics for the STA task.

In the case of EpicKitchen-100 dataset \cite{ek100}, we compare our proposed method against SOTA for \textbf{action anticipation task}, as described in \cite{rulstm,ek55}. It is important to note that \textit{the action anticipation task differs significantly from the STA task}, where the concept of next-active-object is not considered. However, we compute our own annotations to adapt the Action Anticipation task for STA-based scenarios, as discussed in Sec. \ref{sec:datasets}. Since our model and the STA task require the identification of the next-active-object (and its visibility/presence) in the last observed frame, this is reflected in our results due to the limitations of the dataset. The results of our experiments on the EK-100 dataset are presented in Table \ref{tab:ek100_AA}. We achieve state-of-the-art performance on the "Unseen Set" which only contains a small fraction ($6\%$) of samples where no Next-Active-Object (NAO) is detected in the last observed frame. It is to be noted that NAOGAT is the lightest \textit{w.r.t.} other compared models. However, our model's performance on the "Tail Set" is suboptimal, likely due to the fact that the NAO is not visible in the last observed frame for around $22\%$ of the clips. This limitation causes confusion in our model, which relies on the visibility of NAO in the last frame, and impacts the overall results for the "Overall Set," which comprises the "Unseen Set", "Tail Set," and training set's "seen" samples. \\
To investigate the impact of the "Tail Set" on the "Overall Set" accuracy, we remove clips corresponding to tail classes for which NAO is not present in last observed frame and observe improvements of \textbf{$+5.2\uparrow$} (16.9$\%$), \textbf{$+4.0\uparrow$} (32.4$\%$), and \textbf{$+7.6\uparrow$} (35.5$\%$) in action, verb, and noun recognition, respectively.\\
We report additional qualitative results of our model on both dataset in our supplementary material. 

% ## No_futattn_cat - expt name to check the model ckpt

\subsection{Ablation study} 
%In this section, we present the ablation study of our model, where we analyze some of the design choices.
\vspace{-5pt}
%\noindent \textbf{Object and Motion Decoders Ablation. } 
We conducted an ablation study on Ego4D dataset to analyze the impact of different modules of the proposed method in Table \ref{table:ablation}. We evaluated the performance of our complete model in comparison to the models that omit either the Object Decoder (OD) module (Section \ref{subsec:nao}) or the Object Motion Dynamics (OMD) module (Section \ref{subsec:motion} along with Object Decoder (OD) together. Our findings indicate that the Object Decoder module improves the prediction of future verbs ($\hat{v}$), resulting in a higher $AP_{\hat{b} + \hat{v}}$. This suggests that having prior knowledge of the future active-object can support anticipating future action. On the other hand, the OMD module plays a crucial role in estimating the time needed to make contact with the next-active-object and initiate an action, resulting in a significant improvement in $AP_{\hat{b} + \delta}$ and other related metrics. Since OMD provides additional background motion information which helps the model greatly in predicting the TTC. 
These findings suggest that both modules are essential for accurately anticipating future actions in first-person videos. Additionally, we also investigated the impact of the backbone network on our model's performance. For this purpose, we replace our Swin-T backbone with ResNet50 \cite{resnet50} architecture. Using ResNet50 demonstrates a significant drop in performance across all metrics.

% \noindent \textbf{Backbone Ablation.}
% We also conducted the study by replacing the backbone network with a CNN backbone of resnet50 \cite{resnet50} pre-trained on \cite{in1k} for EpicKitchen-100 \cite{ek100} dataset. However, based on the results reported in Table. \ref{tab:ek100_AA_rmvNAO_backboneabs} the model performs poorly for motion-based predictions with the convolution-based backbone, probably due to inability to capture the motion information in the sampled clips. 

\section{Conclusion}
We have investigated the problem of short-term action anticipation using the next-active-objects. First, we discussed the formulation of the STA task. We then presented a new vision transformer-based model, which learns to encode human-object interactions with the help of an object detector and decode the next-active-object location in the last observed frame. We then demonstrated the importance of next-active-object information to predict the future action and time to start the action using additional background information as object motion dynamics.  We proved the proposed method's effectiveness by comparing it against relevant strong anticipation-based baseline methods. 
In future work, we will investigate the use of an object tracker with other human-centered cues such as gaze and the appearance of objects over time. We will also investigate the effect of action recognition on \emph{NAO} identification and localization.

\noindent \textbf{Limitations.} 
As discussed above, the proposed approach is specifically designed for Short-Term Anticipation (STA) task, where the next-active-object is assumed to be visible in the last observed frame. Therefore, when applied to a slightly different task of Action Anticipation, our model shows limitations as it relies on this assumption which does not necessarily hold true in this case. \\
\noindent \textbf{Broad Impact.} The proposed method can be used in several real-world applications such as in robotics or virtual / augmented reality. In case the first person also interacts with other people but not only the non-living objects, then there might be issues regarding privacy preservation. In such cases, policy reviews should be further considered when using the proposed method.   

\clearpage


{\centering\textbf{Leveraging Next-Active Objects for Context-Aware Anticipation in Egocentric Videos: Supplementary Material}} % **** Enter the paper title here
\vspace{30pt}

\maketitle

\begin{figure*}[ht!]
\centering
\includegraphics[width=\linewidth, height=0.35\linewidth]{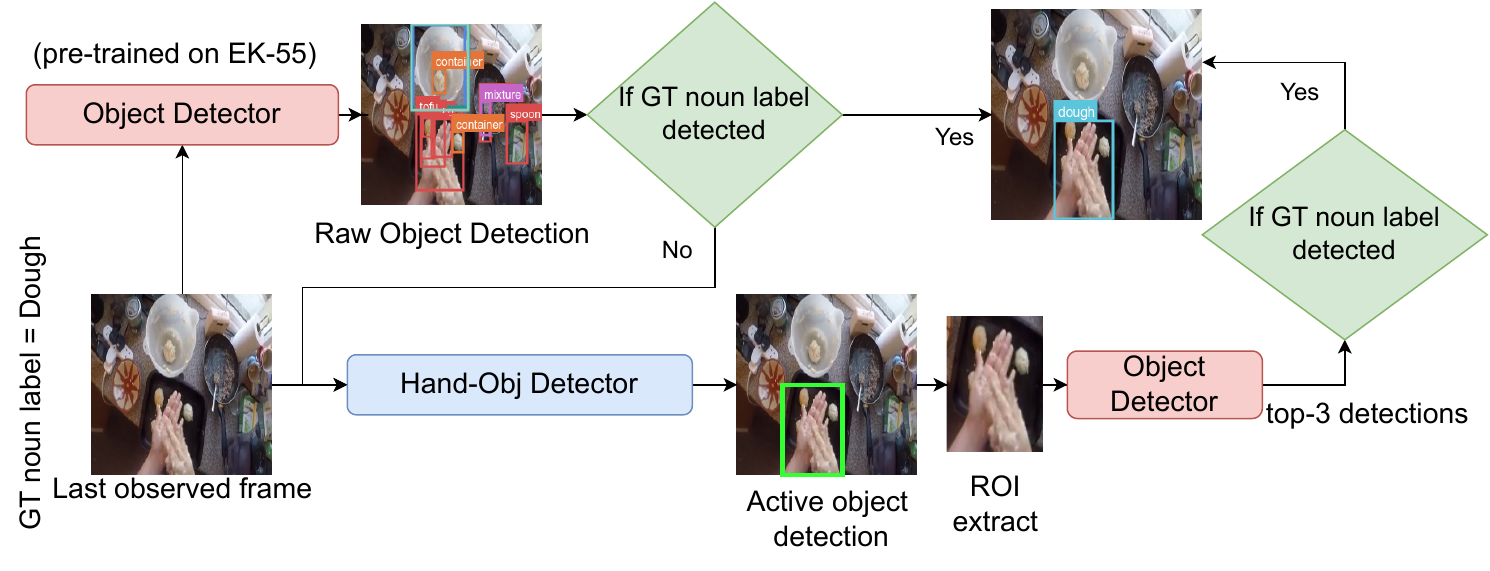}
%\vspace{-0.6cm}
\caption{Annotations pipeline for extracting next-active-object ground-truth labels for EpicKitchen-100 \cite{ek100} dataset. 
} 
\centering
\label{fig:pipeline}
%%%%%%\end{SCfigure*}
\vspace{15pt}
\end{figure*}

\begin{figure*}[ht!]
\centering
\includegraphics[width=\linewidth, height=0.73\linewidth]{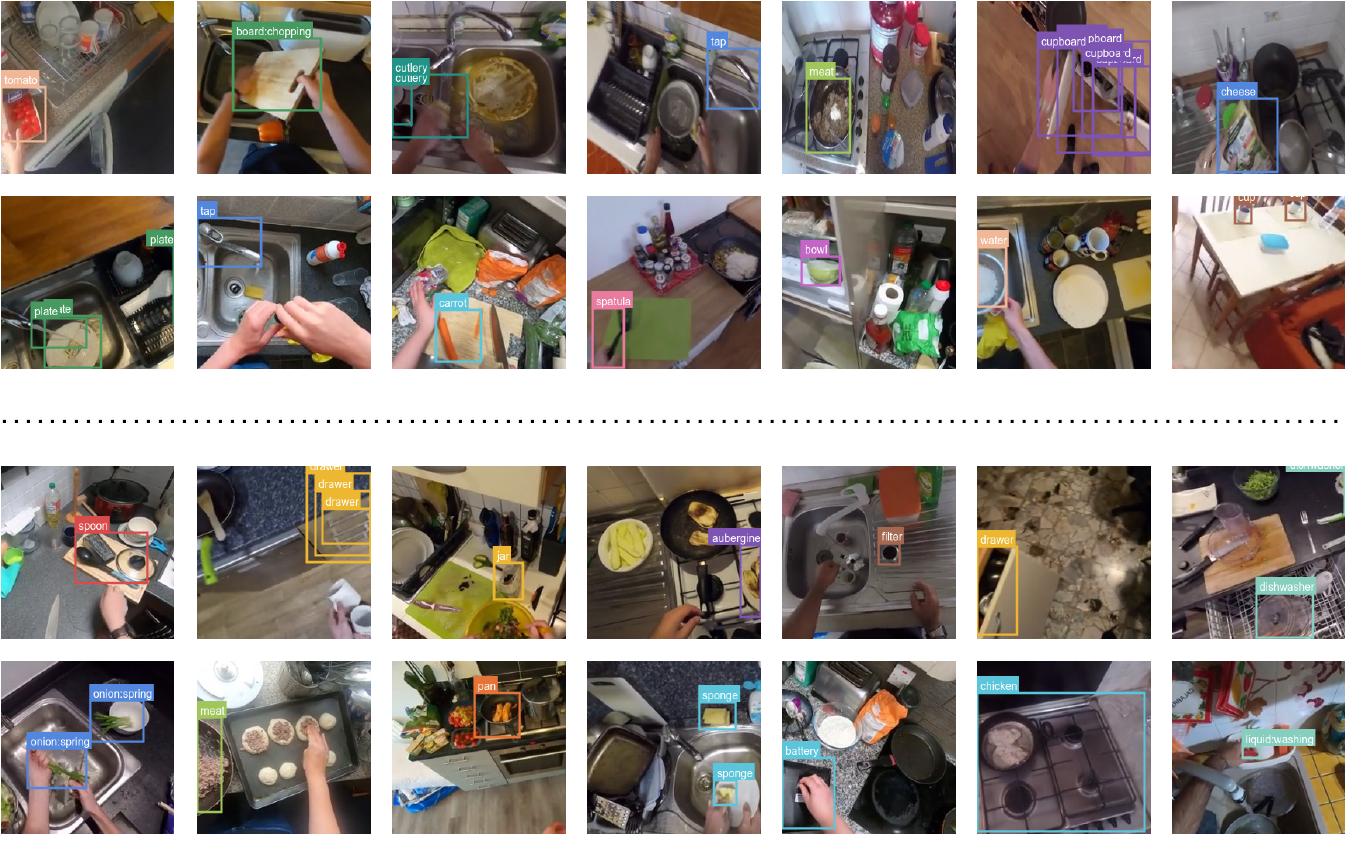}
%\vspace{-0.6cm}
\caption{NAO annotations for EK-100 as curated from the pipeline described in Fig. \ref{fig:pipeline}. The frames corresponds to the last observed frame for a given clip and the detection represents the next-active-object information in terms of NAO location and its class label. 
} 
\centering
\label{fig:ek100_NAO_ann}
%%%%%%\end{SCfigure*}
% \vspace{5pt}
\end{figure*}

This supplementary material presents the qualitative analysis of our model, NAOGAT on Ego4D \cite{ego4d} and EpicKitchen-100 \cite{ek100} dataset. We provide a video depicting the performance of our model when progressed over the allowed the observed segment of a video clip, which is discussed in detail in Sec. \ref{video}. In addition, we also provide some visualization for next-active-object (NAO) annotation on EpicKitchen-100 \cite{ek100}, depicting its location and the class label in the last observed frame for a given video clip. We also describe the annotation pipeline followed to curate the ground-truth data for next-active-object prediction for the Short-Term Anticipation task in Sec. \ref{dataset}.

\section{Video}
\label{video}
We provide additional detail on performance of our model, NAOGAT, when compared with the object detections provided by the object detector pre-trained on Ego4D \cite{ego4d}. We notice a significant improvement in refining the object detections and also identifying objects which are not detected by the object detector to anticipate the location of NAO. The video entails the performance of NAOGAT auto-regressively when fed with a sequential progressive video clip. It can be noticed that as the video progresses, the model further refines the predictions based on past observations and predicts the next-active-object bounding box and its class label, along with future action and time to contact (TTC) with the object.  
The video also provide a visualization on future frames which are not observed by the model describing the time taken to contact with the next-active-object. 

\section{EpicKitchen-100 NAO dataset curation} 
\label{dataset}
The Short-Term Anticipation (STA) task involves predicting the location (bounding box, $\hat{b}$) and class label, $\hat{n}$ of the next-active-object, as well as the future action $\hat{v}$ and the time to contact ($\delta$) with the NAO, for a given video clip. It is important to note that the NAO must be present and visible in the last observed frame for the task to be valid. Currently, only Ego4D \cite{ego4d} dataset provides the precise annotation for studying the problem.  

The EpicKitchen-100 dataset \cite{ek100} offers valuable ground-truth data for the action anticipation \cite{rulstm,avt} task. The dataset includes information on future actions such as "peeling an onion," future verbs like "peel," and associated noun labels of the object involved in the action, such as "onion." This makes the dataset an excellent resource for studying and evaluating models designed to predict future actions. We consider the noun label as the NAO class label for a given clip. However, it lacks annotations for the location of NAO in the last observed frame. For this purpose, we aimed to curate our own annotation for NAO estimation following the pipeline described in Fig. \ref{fig:pipeline}. 

To curate ground-truth data for the next-active-object prediction for the Short-Term Anticipation task, we first extract the last observed frame from a given clip. Next, we use a pre-trained object detector \cite{fastercnn} on the EK-55 dataset \cite{ek55} to obtain raw object detections for the frame. We then verify if the ground-truth NAO class label is identified in the raw detections. If a match is found, the corresponding bounding box for that detection is used as the ground-truth annotation for the NAO bounding box ($\hat{b}$).
However, if the object detector fails to identify any object with the ground-truth NAO label, we use a Hand-Object detector \cite{Shan20} to obtain bounding boxes for the active object \cite{ADL}. This is because the hand-object detector has been shown to be state-of-the-art in identifying hand-object detection and has been used in the literature \cite{anacto,hand_obj_joint}. In the event that the Hand-Object detector identifies an active object, we extract the Region of Interest (ROI) for the corresponding detection from the input frame. This ROI is then fed into the object detector \cite{fastercnn} used earlier, and we take the top-3 predictions from the detector. These predictions are once again verified against the Ground-Truth NAO class label to check if they contain the NAO label. If one of the predictions satisfies the criteria, the location of the active object is used as the ground-truth annotation for the NAO location. This pipeline is used to only curate information regarding the location of NAO and not the class label of NAO for a given clip. The class label for NAO is used from the annotations provided with EK-100 for action anticipation. The final annotations for the dataset are shown in Fig. \ref{fig:ek100_NAO_ann}.

\begin{figure*}[ht!]
\centering
\includegraphics[width=\linewidth, height=0.60\linewidth]{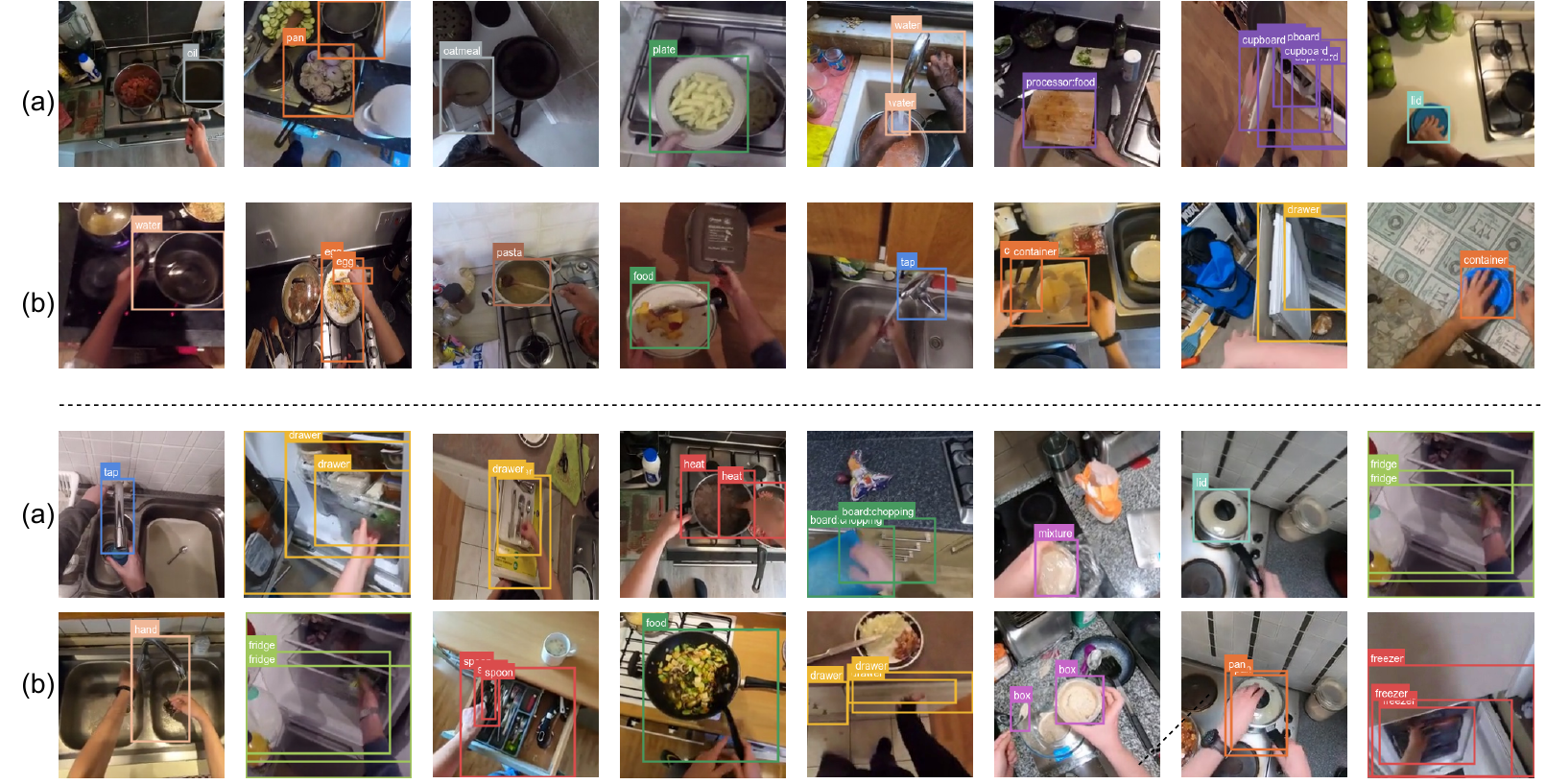}
%\vspace{-0.6cm}
\caption{Due to the large number of noun labels in EK-100, similar-looking objects are labeled differently multiple times in the dataset. This confuses our model, NAOGAT since the future action prediction is affected based on the NAO prediction. 
} 
\centering
\label{fig:ek100_multiple_obj}
%%%%%%\end{SCfigure*}
% \vspace{-5pt}
\end{figure*}

\begin{figure*}[ht!]
\centering
\includegraphics[width=\linewidth, height=0.30\linewidth]{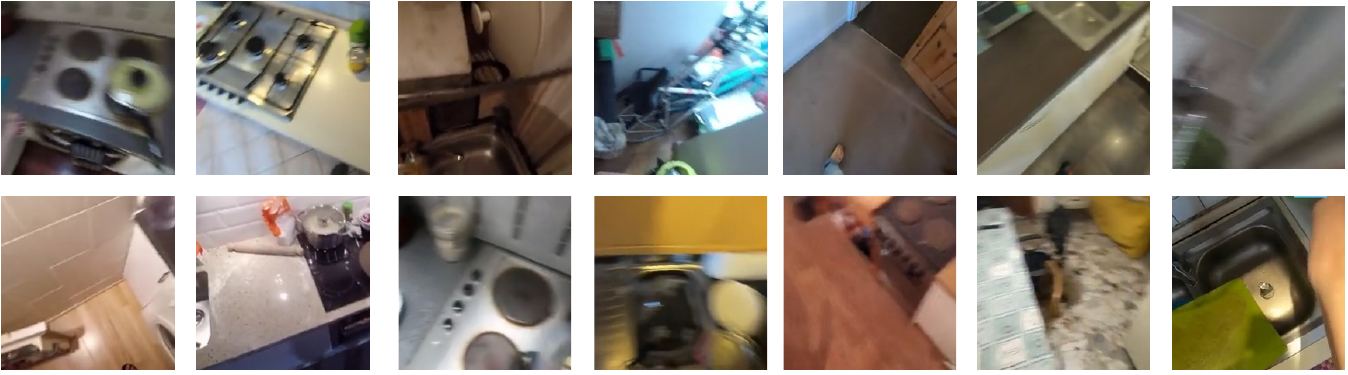}
%\vspace{-0.6cm}
\caption{Instances in EpicKitchen-100 where the next-active-object is not detected / not present in the last observed frame.
} 
\centering
\label{fig:ek100_fail_ann}
%%%%%%\end{SCfigure*}
% \vspace{-5pt}
\end{figure*}
%%%%%%%%% BODY TEXT
\section{Limitations of our model for EpicKitchen-100 dataset}
It is important to note that EpicKitchen-100 was not curated in alignment with the definition of STA. Specifically, the dataset does not provide annotations for next-active-object, and it is not mandatory for NAO to be present in the allowed last frame observed by the model. As discussed in the main paper, our dataset curation method (described in Sec. \ref{dataset}) could not annotate the ground-truth data for the next-active-object in \emph{22\%} of the "Test Set" of the Validation split, as there were no detected objects in those clips. Moreover, the EK-100 dataset suffers from a dataset bias, as there are \emph{300} class labels for objects, and similar-looking objects are often classified differently, as shown in Fig. \ref{fig:ek100_multiple_obj}. This further confuses the model's identification of objects and impedes its ability to anticipate future actions.

{\small
\bibliographystyle{ieee_fullname}
\bibliography{main}

\begin{thebibliography}{10}\itemsep=-1pt

\bibitem{borghi2005object}
Anna~M Borghi.
\newblock Object concepts and action.
\newblock {\em Grounding cognition: The role of perception and action in
  memory, language, and thinking}, pages 8--34, 2005.

\bibitem{detr}
Nicolas Carion, Francisco Massa, Gabriel Synnaeve, Nicolas Usunier, Alexander
  Kirillov, and Sergey Zagoruyko.
\newblock End-to-end object detection with transformers.
\newblock In Andrea Vedaldi, Horst Bischof, Thomas Brox, and Jan-Michael Frahm,
  editors, {\em Computer Vision -- ECCV 2020}, pages 213--229, Cham, 2020.
  Springer International Publishing.

\bibitem{i3d}
J. Carreira and A. Zisserman.
\newblock Quo vadis, action recognition? a new model and the kinetics dataset.
\newblock In {\em 2017 IEEE Conference on Computer Vision and Pattern
  Recognition (CVPR)}, pages 4724--4733, Los Alamitos, CA, USA, jul 2017. IEEE
  Computer Society.

\bibitem{ek100}
Dima Damen, Hazel Doughty, Giovanni~Maria Farinella, , Antonino Furnari, Jian
  Ma, Evangelos Kazakos, Davide Moltisanti, Jonathan Munro, Toby Perrett, Will
  Price, and Michael Wray.
\newblock Rescaling egocentric vision.
\newblock {\em International Journal of Computer Vision}, 2021.

\bibitem{ek55}
Dima Damen, Hazel Doughty, Giovanni~Maria Farinella, Sanja Fidler, Antonino
  Furnari, Evangelos Kazakos, Davide Moltisanti, Jonathan Munro, Toby Perrett,
  Will Price, and Michael Wray.
\newblock Scaling egocentric vision: The epic-kitchens dataset.
\newblock In {\em European Conference on Computer Vision (ECCV)}, 2018.

\bibitem{in1k}
Jia Deng, Wei Dong, Richard Socher, Li-Jia Li, Kai Li, and Li Fei-Fei.
\newblock Imagenet: A large-scale hierarchical image database.
\newblock In {\em 2009 IEEE Conference on Computer Vision and Pattern
  Recognition}, pages 248--255, 2009.

\bibitem{tpami_contact}
Eadom Dessalene, Chinmaya Devaraj, Michael Maynord, Cornelia Fermuller, and
  Yiannis Aloimonos.
\newblock Forecasting action through contact representations from first person
  video.
\newblock {\em IEEE TPAMI}, pages 1--1, 2021.

\bibitem{ego_obj_graph}
Eadom Dessalene, Michael Maynord, Chinmaya Devaraj, Cornelia Fermuller, and
  Yiannis Aloimonos.
\newblock Egocentric object manipulation graphs.
\newblock {\em arXiv preprint arXiv:2006.03201}, 2020.

\bibitem{vit}
Alexey Dosovitskiy, Lucas Beyer, Alexander Kolesnikov, Dirk Weissenborn,
  Xiaohua Zhai, Thomas Unterthiner, Mostafa Dehghani, Matthias Minderer, Georg
  Heigold, Sylvain Gelly, Jakob Uszkoreit, and Neil Houlsby.
\newblock An image is worth 16x16 words: Transformers for image recognition at
  scale.
\newblock In {\em International Conference on Learning Representations}, 2021.

\bibitem{feichtenhofer2019slowfast}
Christoph Feichtenhofer, Haoqi Fan, Jitendra Malik, and Kaiming He.
\newblock Slowfast networks for video recognition.
\newblock In {\em Proceedings of the IEEE/CVF international conference on
  computer vision}, pages 6202--6211, 2019.

\bibitem{furnari2017next}
Antonino Furnari, Sebastiano Battiato, Kristen Grauman, and Giovanni~Maria
  Farinella.
\newblock Next-active-object prediction from egocentric videos.
\newblock {\em Journal of Visual Communication and Image Representation},
  49:401--411, 2017.

\bibitem{rulstm}
Antonino Furnari and Giovanni~Maria Farinella.
\newblock What would you expect? anticipating egocentric actions with
  rolling-unrolling lstms and modality attention.
\newblock In {\em International Conference on Computer Vision}, 2019.

\bibitem{avt}
Rohit Girdhar and Kristen Grauman.
\newblock {Anticipative Video Transformer}.
\newblock In {\em ICCV}, 2021.

\bibitem{smoothl1}
Ross Girshick.
\newblock Fast r-cnn.
\newblock In {\em IEEE ICCV}, pages 1440--1448, 2015.

\bibitem{ego4d}
Kristen Grauman, Andrew Westbury, and Eugene et~al. Byrne.
\newblock Ego4d: Around the {W}orld in 3,000 {H}ours of {E}gocentric {V}ideo.
\newblock In {\em IEEE/CVF Computer Vision and Pattern Recognition (CVPR)},
  2022.

\bibitem{resnet50}
Kaiming He, Xiangyu Zhang, Shaoqing Ren, and Jian Sun.
\newblock Deep residual learning for image recognition.
\newblock In {\em Proceedings of the IEEE Conference on Computer Vision and
  Pattern Recognition (CVPR)}, June 2016.

\bibitem{ar_man}
Julia Hertel, Sukran Karaosmanoglu, Susanne Schmidt, Julia Bräker, Martin
  Semmann, and Frank Steinicke.
\newblock A taxonomy of interaction techniques for immersive augmented reality
  based on an iterative literature review.
\newblock In {\em 2021 IEEE International Symposium on Mixed and Augmented
  Reality (ISMAR)}, pages 431--440, 2021.

\bibitem{orvit}
Roei Herzig, Elad Ben-Avraham, Karttikeya Mangalam, Amir Bar, Gal Chechik, Anna
  Rohrbach, Trevor Darrell, and Amir Globerson.
\newblock Object-region video transformers.
\newblock In {\em Proceedings of the IEEE/CVF Conference on Computer Vision and
  Pattern Recognition (CVPR)}, pages 3148--3159, June 2022.

\bibitem{MD_1}
Max Jaderberg, Karen Simonyan, Andrew Zisserman, and koray kavukcuoglu.
\newblock Spatial transformer networks.
\newblock In C. Cortes, N. Lawrence, D. Lee, M. Sugiyama, and R. Garnett,
  editors, {\em Advances in Neural Information Processing Systems}, volume~28.
  Curran Associates, Inc., 2015.

\bibitem{JIANG2021212}
Jingjing Jiang, Zhixiong Nan, Hui Chen, Shitao Chen, and Nanning Zheng.
\newblock Predicting short-term next-active-object through visual attention and
  hand position.
\newblock {\em Neurocomputing}, 433:212--222, 2021.

\bibitem{k400}
Will Kay, Joao Carreira, Karen Simonyan, Brian Zhang, Chloe Hillier, Sudheendra
  Vijayanarasimhan, Fabio Viola, Tim Green, Trevor Back, Paul Natsev, Mustafa
  Suleyman, and Andrew Zisserman.
\newblock The kinetics human action video dataset, 2017.

\bibitem{obj_handobj}
Kyungjun Lee, Abhinav Shrivastava, and Hernisa Kacorri.
\newblock Leveraging hand-object interactions in assistive egocentric vision.
\newblock {\em IEEE Transactions on Pattern Analysis and Machine Intelligence},
  pages 1--1, 2021.

\bibitem{liu2019forecasting}
Miao Liu, Siyu Tang, Yin Li, and James Rehg.
\newblock Forecasting human object interaction: Joint prediction of motor
  attention and actions in first person video.
\newblock In {\em ECCV}, 2020.

\bibitem{hand_obj_joint}
Shaowei Liu, Subarna Tripathi, Somdeb Majumdar, and Xiaolong Wang.
\newblock Joint hand motion and interaction hotspots prediction from egocentric
  videos.
\newblock In {\em Proceedings of the IEEE/CVF Conference on Computer Vision and
  Pattern Recognition (CVPR)}, 2022.

\bibitem{swin}
Ze Liu, Yutong Lin, Yue Cao, Han Hu, Yixuan Wei, Zheng Zhang, Stephen Lin, and
  Baining Guo.
\newblock Swin transformer: Hierarchical vision transformer using shifted
  windows.
\newblock In {\em Proceedings of the IEEE/CVF International Conference on
  Computer Vision (ICCV)}, 2021.

\bibitem{swin2}
Ze Liu, Yutong Lin, Yue Cao, Han Hu, Yixuan Wei, Zheng Zhang, Stephen Lin, and
  Baining Guo.
\newblock Swin transformer: Hierarchical vision transformer using shifted
  windows.
\newblock {\em arXiv preprint arXiv:2103.14030}, 2021.

\bibitem{swin1}
Ze Liu, Jia Ning, Yue Cao, Yixuan Wei, Zheng Zhang, Stephen Lin, and Han Hu.
\newblock Video swin transformer.
\newblock {\em arXiv preprint arXiv:2106.13230}, 2021.

\bibitem{video_swin}
Ze Liu, Jia Ning, Yue Cao, Yixuan Wei, Zheng Zhang, Stephen Lin, and Han Hu.
\newblock Video swin transformer.
\newblock In {\em Proceedings of the IEEE/CVF Conference on Computer Vision and
  Pattern Recognition (CVPR)}, pages 3202--3211, June 2022.

\bibitem{vr_obj_man}
Blascovich~J.J. Loomis~J.M. and A.C. Beall.
\newblock Immersive virtual environment technology as a basic research tool in
  psychology.
\newblock {\em Behavior Research Methods, Instruments, and Computers},
  31:557–564, 1999.

\bibitem{obj1}
Joanna Materzynska, Tete Xiao, Roei Herzig, Huijuan Xu, Xiaolong Wang, and
  Trevor Darrell.
\newblock Something-else: Compositional action recognition with
  spatial-temporal interaction networks.
\newblock In {\em 2020 IEEE/CVF Conference on Computer Vision and Pattern
  Recognition (CVPR)}, pages 1046--1056, 2020.

\bibitem{obj_emb_behaviour}
Tushar Nagarajan and Kristen Grauman.
\newblock Shaping embodied agent behavior with activity-context priors from
  egocentric video.
\newblock In M. Ranzato, A. Beygelzimer, Y. Dauphin, P.S. Liang, and J.~Wortman
  Vaughan, editors, {\em Advances in Neural Information Processing Systems},
  volume~34, pages 29794--29805. Curran Associates, Inc., 2021.

\bibitem{ADL}
Hamed Pirsiavash and Deva Ramanan.
\newblock Detecting activities of daily living in first-person camera views.
\newblock In {\em IEEE CVPR}, pages 2847--2854, 2012.

\bibitem{gpt2}
Alec Radford, Jeff Wu, Rewon Child, David Luan, Dario Amodei, and Ilya
  Sutskever.
\newblock Language models are unsupervised multitask learners.
\newblock 2019.

\bibitem{meccano}
Francesco Ragusa, Antonino Furnari, Salvatore Livatino, and Giovanni~Maria
  Farinella.
\newblock The meccano dataset: Understanding human-object interactions from
  egocentric videos in an industrial-like domain.
\newblock In {\em Proceedings of the IEEE/CVF Winter Conference on Applications
  of Computer Vision (WACV)}, pages 1569--1578, January 2021.

\bibitem{fastercnn}
Shaoqing Ren, Kaiming He, Ross Girshick, and Jian Sun.
\newblock Faster r-cnn: Towards real-time object detection with region proposal
  networks.
\newblock In C. Cortes, N. Lawrence, D. Lee, M. Sugiyama, and R. Garnett,
  editors, {\em Advances in Neural Information Processing Systems}, volume~28.
  Curran Associates, Inc., 2015.

\bibitem{sener2020temporal}
Fadime Sener, Dipika Singhania, and Angela Yao.
\newblock Temporal aggregate representations for long-range video
  understanding.
\newblock In {\em European Conference on Computer Vision}, pages 154--171.
  Springer, 2020.

\bibitem{Shan20}
Dandan Shan, Jiaqi Geng, Michelle Shu, and David Fouhey.
\newblock Understanding human hands in contact at internet scale.
\newblock In {\em Proceedings of the IEEE Conference on Computer Vision and
  Pattern Recognition (CVPR)}, 2020.

\bibitem{anacto}
Sanket Thakur, Cigdem Beyan, Pietro Morerio, Vittorio Murino, and Alessio
  Del~Bue.
\newblock Anticipating next active objects for egocentric videos, 2023.

\bibitem{obj_AR}
Xiaohan Wang, Linchao Zhu, Heng Wang, and Yi Yang.
\newblock Interactive prototype learning for egocentric action recognition.
\newblock In {\em Proceedings of the IEEE/CVF International Conference on
  Computer Vision (ICCV)}, pages 8168--8177, October 2021.

\bibitem{memvit2022}
Chao-Yuan Wu, Yanghao Li, Karttikeya Mangalam, Haoqi Fan, Bo Xiong, Jitendra
  Malik, and Christoph Feichtenhofer.
\newblock {MeMViT: Memory-Augmented Multiscale Vision Transformer for Efficient
  Long-Term Video Recognition}.
\newblock In {\em CVPR}, 2022.

\bibitem{future_frame}
Yu Wu, Linchao Zhu, Xiaohan Wang, Yi Yang, and Fei Wu.
\newblock Learning to anticipate egocentric actions by imagination.
\newblock {\em IEEE Transactions on Image Processing}, 30:1143--1152, 2020.

\bibitem{Zhong_2023_WACV}
Zeyun Zhong, David Schneider, Michael Voit, Rainer Stiefelhagen, and J\"urgen
  Beyerer.
\newblock Anticipative feature fusion transformer for multi-modal action
  anticipation.
\newblock In {\em Proceedings of the IEEE/CVF Winter Conference on Applications
  of Computer Vision (WACV)}, pages 6068--6077, January 2023.

\end{thebibliography}
}

\end{document}